\documentclass[10pt,twocolumn,letterpaper]{article}
\usepackage{cvpr}              % To produce the CAMERA-READY version
\definecolor{cvprblue}{rgb}{0.21,0.49,0.74}
\usepackage[pagebackref,breaklinks,colorlinks,allcolors=cvprblue]{hyperref}
\usepackage{algorithm}
\usepackage{algpseudocode}
\usepackage{multirow}
\usepackage{pifont}
\usepackage{colortbl}
\usepackage{graphicx}
\usepackage{float}
\usepackage{caption}
\usepackage{subcaption}
\def\paperID{9941} % *** Enter the Paper ID here
\def\confName{CVPR}
\def\confYear{2025}
\hypersetup{
    colorlinks=true, 
    urlcolor=magenta,   
}
\title{From Poses to Identity: Training-Free Person Re-Identification \\ via Feature Centralization}

\author{
Chao Yuan$^{1,*}$, Guiwei Zhang$^{1,*}$, Changxiao Ma$^{1}$, Tianyi Zhang$^{2}$, Guanglin Niu$^{2,\dagger}$ \\
\textsuperscript{1} School of Computer Science and Engineering, Beihang University \quad \\
\textsuperscript{2} School of Artificial Intelligence, Beihang University  \\
{\tt\small yuanc3666@gmail.com}, {\tt\small \{zhangguiwei,cxma124,zy2442222,beihangngl\}@buaa.edu.cn} \\
}

\begin{document}

\twocolumn[{%
\renewcommand\twocolumn[1][]{#1}%
\maketitle
\begin{center}
    \centering
    \includegraphics[width=0.8\textwidth]{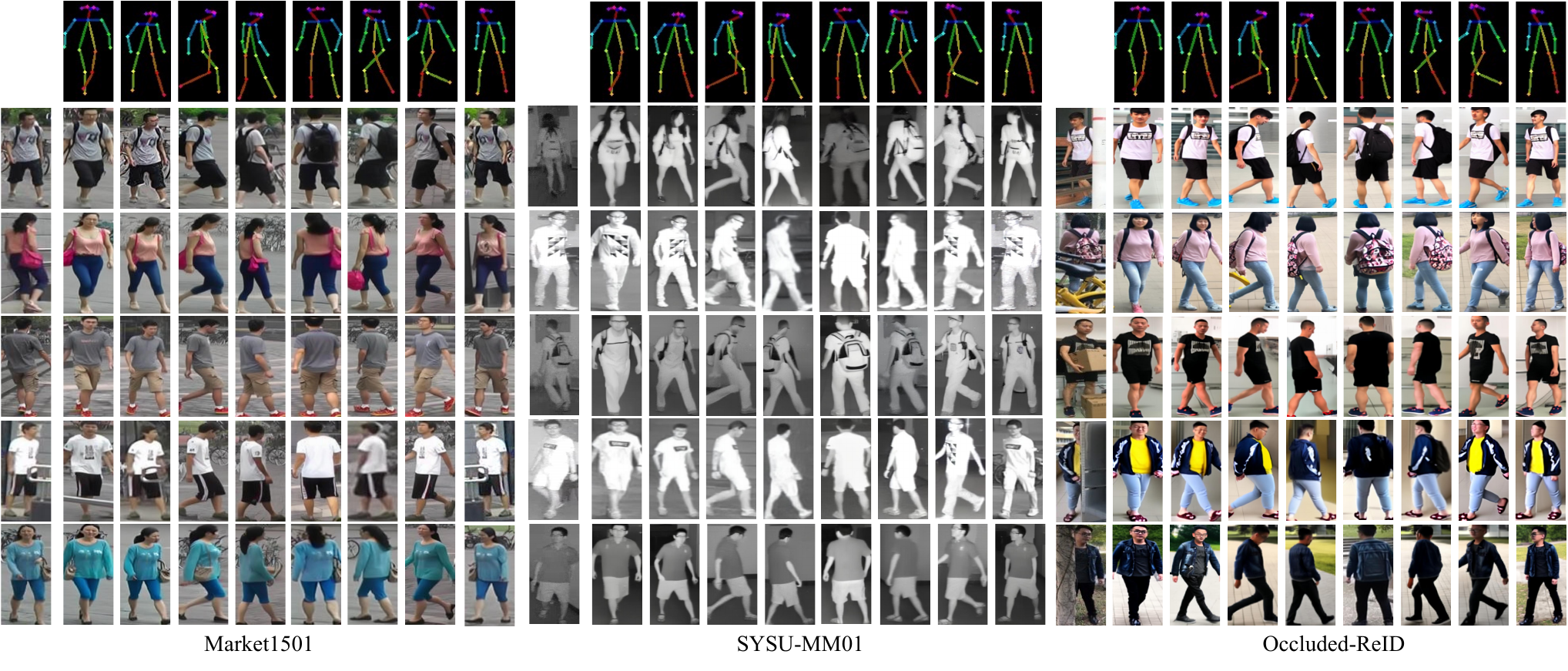} 
    \captionof{figure}{Visualization of our Identity-Guided Pedestrian Generation model with 8 representative poses on three datasets.}
\end{center}%
}]
\renewcommand{\thefootnote}{}
\footnotetext{$\dagger$ Corresponding author. * Equal contribution.}
\footnotetext{Codes: \textit{\url{https://github.com/yuanc3/Pose2ID}}}
% \maketitle

% \begin{figure*}[t]
%     \centering
%     \includegraphics[width=0.9\textwidth]{figs/pdf/visualization.pdf} 
%     \label{fig:visualization}
%     \caption{Visualization of our Pedestrian Generation model with 8 directions 'standard poses' on Market1501, SYSU-MM01, and Occluded-ReID datasets.}
% \end{figure*}

\begin{abstract}
Person re-identification (ReID) aims to extract accurate identity representation features. However, during feature extraction, individual samples are inevitably affected by noise (background, occlusions, and model limitations). Considering that features from the same identity follow a normal distribution around identity centers after training, we propose a \textbf{Training-Free} \textbf{Feature Centralization} ReID framework (\textbf{Pose2ID}) by aggregating the same identity features to reduce individual noise and enhance the stability of identity representation, which preserves the feature's original distribution for following strategies such as re-ranking. Specifically, to obtain samples of the same identity, we introduce two components:
\ding{192}Identity-Guided Pedestrian Generation: by leveraging identity features to guide the generation process, we obtain high-quality images with diverse poses, ensuring identity consistency even in complex scenarios such as infrared, and occlusion.
\ding{193}Neighbor Feature Centralization: it explores each sample's potential positive samples from its neighborhood.
Experiments demonstrate that our generative model exhibits strong generalization capabilities and maintains high identity consistency. With the Feature Centralization framework, we achieve impressive performance even with an ImageNet pre-trained model without ReID training, reaching mAP/Rank-1 of 52.81/78.92 on Market1501. Moreover, our method sets new state-of-the-art results across standard, cross-modality, and occluded ReID tasks, showcasing strong adaptability.
\end{abstract}    
\section{Introduction}
\label{sec:intro}

Person re-identification (ReID) is a critical task in video surveillance and security, aiming to match pedestrian images captured from different cameras. Despite significant progress made in recent years through designing more complex models with increased parameters and feature dimensions, inevitable noise arises due to poor image quality or inherent limitations of the models, reducing the accuracy of identity recognition and affecting retrieval performance.

\begin{figure}
\centering
\includegraphics[width=0.85\linewidth]{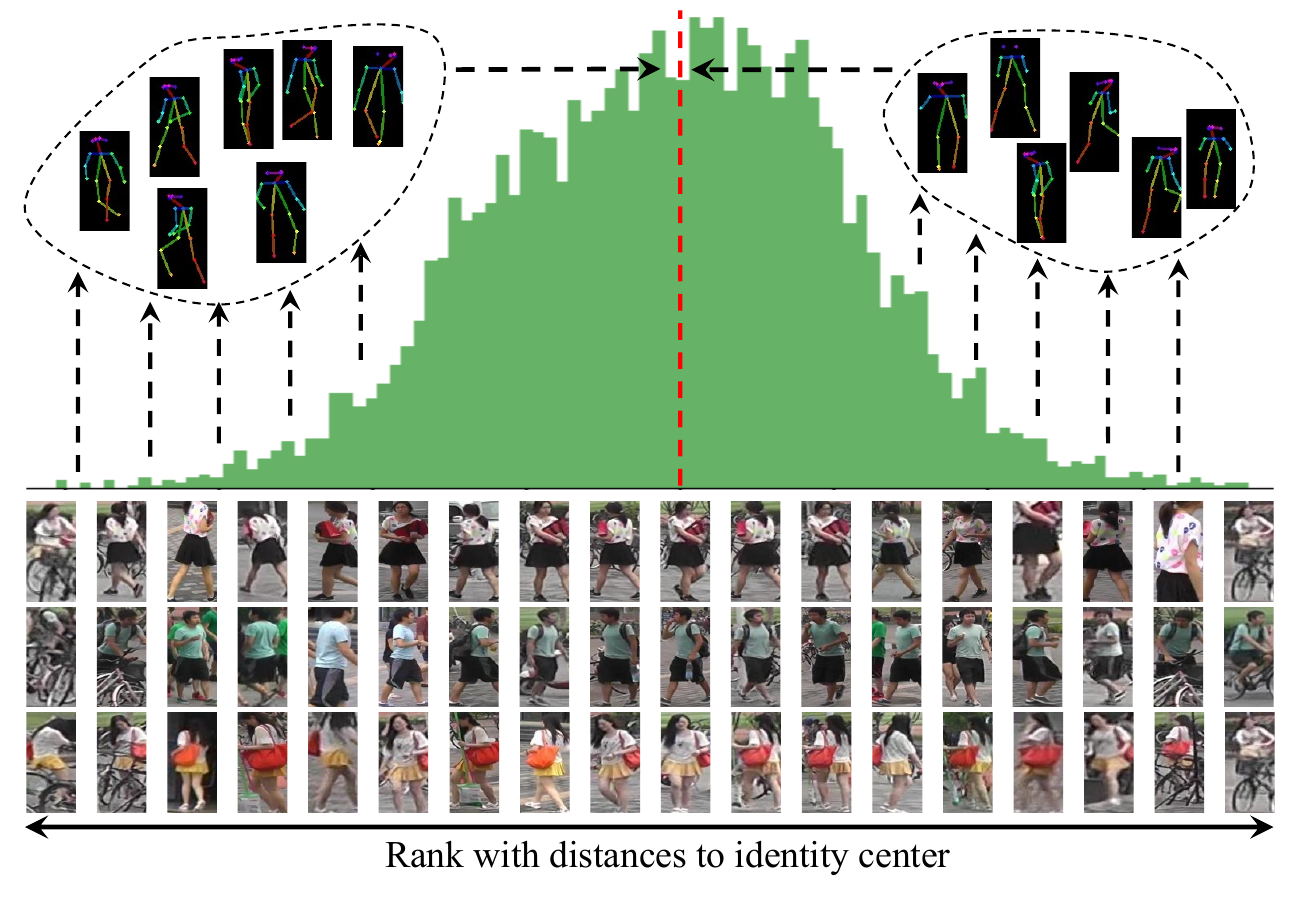}
\caption{The real feature distribution of images from the same ID extracted by TransReID, and the main idea of our work is to make features closer/centralized to the ID center.}
\label{fig:intro}
\end{figure}

To address this challenge, we propose a training-free ReID framework that fully leverages capabilities of existing models by mitigating feature noise to enhance identity representation. During training, features of the same identity are constrained by the loss function and naturally aggregate around an "identity center" in the feature space. Particularly, according to the central limit theorem, when the number of samples is sufficiently large, these features would follow a normal distribution with identity center as the mean. As shown in Fig.~\ref{fig:intro}, we visualize the feature distribution of an identity, and rank samples of the same identity with distance to their identity center. Therefore, we introduce the concept of feature centralization. To make each sample's feature more representative of its identity, by aggregating features of the same identity, we can reduce the noise in individual samples, strengthen identity characteristics, and bring each feature dimension closer to its central value. 

However, obtaining diverse samples of the same identity is hard without identity labels. With the development of generative models, generating images of the same identity in different poses has become feasible. Previous GAN-based studies \cite{ge2018fd, zheng2017unlabeled, zheng2019joint, qian2018pose} struggle with limited effectiveness, and mainly serve to augment training data. With breakthroughs in diffusion models for image generation \cite{ho2020denoising, alexey2020image, zhang2023adding}, it is now possible to generate high-quality, multi-pose images of the same person. However, these methods lack effective control over identity features and generated pedestrian images are susceptible to interference from background or occlusions, making it difficult to ensure identity consistency across poses. Therefore, we utilize identity features of ReID, proposing an Identity-Guided Pedestrian Generation paradigm. Guided by identity features, we generate high-quality images of the same person with a high degree of identity consistency across diverse scenarios (visible, infrared, and occluded).

Furthermore, inspired by re-ranking mechanisms\cite{zhong2017re}, we explore potential positive samples through feature distance matrices to further achieve feature centralization. Unlike traditional re-ranking methods that modify features or distances in a one-off manner, our approach performs L2 normalization after enhancement. This preserves the original feature distribution while improving representation quality and can be combined with re-ranking methods.

Thus, the main contributions of this paper include:
\begin{itemize}
    \item[$\bullet$]\textbf{Training-Free} \textbf{Feature Centralization} framework that can be directly applied to different ReID tasks/models, even an ImageNet pretrained ViT without ReID training;
    \item[$\bullet$]\textbf{I}dentity-Guided \textbf{P}edestrian \textbf{G}eneration (IPG) paradigm, leveraging identity features to generate high-quality images of the same identity in different poses to achieve feature centralization;
    \item[$\bullet$]\textbf{N}eighbor \textbf{F}eature \textbf{C}entralization (NFC) based on sample's neighborhood, discovering hidden positive samples from gallery/query set to achieve feature centralization.
\end{itemize}

% 行人重识别（Person Re-identification, ReID）任务是一个图像检索问题，目标是从行人图像中提取行人特征，并利用这些特征进行精确匹配与检索。当前的大部分研究集中在构建强大的特征提取模型上，通常通过增加模型参数和特征维度以便从有限数据中学习区分性特征。然而，在实际应用中，因图像质量或模型局限性，部分提取到的特征可能不属于行人身份本身，反而成为噪声特征。现有方法通常未关注噪声特征的存在，无法有效应对其影响。因此，如何有效抑制特征中的噪声并增强其身份表示能力，成为本文研究的关键问题。

% 特征噪声虽难以完全消除，但可以通过合适的方法予以削弱。基于此，我们提出了特征“身份中心化”的概念。得益于ReID模型的损失函数，训练过程自然会促使同一身份的样本特征逐渐收敛于特征空间中的一个聚集点，我们称之为“身份中心”。根据中心极限定理，当样本量足够大时，同一身份的特征会围绕这个隐式的身份中心呈正态分布。由此，可以通过叠加同一身份的特征，使得单个特征更接近其身份中心，从而弱化噪声并增强其身份代表性。

% 然而，如何获取相同身份的样本是一个挑战，因为我们通常无法直接获取每个样本的身份标签。最简单的就是水平反转来获得一个额外特征，这也是ReID领域中很多热你会用到的一个trick。此外，受到Re-rank机制的启发，我们可以通过特征的距离矩阵去挖掘潜在的正样本对特征进行中心化。但Re-rank是一种一次性的方法，通过加权，缩放等去修改特征或者距离矩阵，而我们进行特征增强后，会进行L2 norm来归一化，维持了特征的原有的分布。

% 当然，随着生成模型的快速发展，生成同一身份的不同姿态图像成为可能。例如，ABC等人尝试了基于GAN的生成方法，但生成效果有限，主要用于扩增训练数据。随着扩散模型在图像生成领域取得突破，CDF等人验证了生成同一人不同姿态高质量图像的可行性。然而，这些方法缺乏身份特征的指导，在行人图像生成中往往受到背景信息和遮挡物的干扰，难以保证不同姿态生成图像的身份一致性。

% 为了解决上述问题，本文提出了一种无需额外训练的行人ReID框架。本文的主要贡献包括：

% 详细分析了ReID特征的分布特性，提出了一种模型无关的特征增强框架，即使基于预训练的ViT模型也能直接应用于ReID任务；
% 提出了一个基于身份特征指导的行人图像生成框架，利用身份信息生成同一身份不同姿态的高质量图像，以实现特征增强；
% 设计了一种基于互惠信息挖掘的特征增强算法，从样本集中高效挖掘潜在信息以提升特征的身份表示能力。

% zheng等人的文章指出，即使图片质量再差，甚至不是一个“人”，其身份的预测最大概率仍然会落在其对应的identity上。在特征维度，任意两个样本，只要他们是来自同一个身份个体的，他们的特征空间总会有关联。

%-------------------------------------------------------------------------

\section{Related works}
\paragraph{Person Re-Identification}
Person re-identification (ReID) is a critical task in computer vision that focuses on identifying individuals across different camera views. It plays a significant role in surveillance and security applications. Recent advancements in ReID have leveraged deep learning techniques to enhance performance, particularly using convolutional neural networks (CNNs\cite{krizhevsky2012imagenet}) and vision transformers (ViTs \cite{alexey2020image}). The deep learning based methods\cite{bak2010person, layne2012person, sun2017svdnet, hermans2017defense, liao2020interpretable, fu2021unsupervised, zhang2023protohpe, zhang2023pha} that focus on feature extraction and metric learning\cite{koestinger2012large, hirzer2012relaxed, liao2015efficient, yu2018unsupervised} have improved feature extraction by learning robust and discriminative embeddings that capture identity-specific information.

Standard datasets like Market-1501 \cite{zheng2015scalable} have been widely used to benchmark ReID algorithms under normal conditions. Moreover, there are some challenging scenarios such as occlusions and cross-modality matching. Occluded-REID \cite{zhuo2018occluded} addresses the difficulties of identifying partially obscured individuals, while SYSU-MM01 \cite{wu2017rgb} focuses on matching identities between visible and infrared images, crucial for nighttime surveillance.

\paragraph{Feature Enhancement in Re-Identification}
Extracting robust feature representations is one of the key challenges in re-identification. Feature enhancement could help the ReID model easily differentiate between two people. Data augmentation techniques\cite{mclaughlin2015data, zhong2020random, ma2019true} were enhanced for feature enhancement. By increasing the diversity of training data, ReID model could extract robust and discriminative features.

Apart from improving the quality of the generated images, some Gan-based methods couple feature extraction and data generation end-to-end to distill identity related feature. FD-GAN\cite{ge2018fd} separates identity from the pose by generating images across multiple poses, enhancing the ReID system's robustness to pose changes. Zheng et al.\cite{zheng2019joint} separately encodes each person into an appearance code and a structure code. Eom el al.\cite{eom2019learning} propose to disentangle identity-related and identity-unrelated features from person images.  However, GAN-based methods face challenges such as training instability and mode collapse, which may not keep identity consistency.

In addition, the re-ranking technique refines feature-based distances to improve ReID accuracy. Such as k-Reciprocal Encoding Re-ranking\cite{zhong2017re}, which improves retrieval accuracy by leveraging the mutual neighbors between images to refine the distance metric. 

\paragraph{Person Generation Models}
Recent approaches have incorporated generative models, particularly generative adversarial networks (GANs) \cite{goodfellow2014generative}, to augment ReID data or enhance feature quality. Style transfer GAN-based methods\cite{zhong2018camstyle, dai2018cross, huang2019sbsgan, pang2022cross} transfer labeled training images to artificially generated images in diffrent camera domains, background domains or RGB-infrared domains. Pose-transfer GAN-based methods\cite{siarohin2018deformable,qian2018pose, liu2018pose, borgia2019gan, zhang2020pac} enable the synthesis of person images with variations in pose and appearance, enriching the dataset and making feature representations more robust to changes in poses. Random generation GAN-based mothods \cite{zheng2017unlabeled, ainam2019sparse, hussin2021stylegan} generate random images of persons and use Label Smooth Regularization (LSR \cite{szegedy2016rethinking}) or other methods to automatically label them. However, these methods often struggle to maintain identity consistency in pose variation, as generated images are susceptible to identity drift. 
The emergence of diffusion models has advanced the field of generative modeling, showing remarkable results in image generation tasks \cite{ho2020denoising}. Leveraging the capabilities of pre-trained models like Stable Diffusion \cite{rombach2022high}, researchers have developed techniques\cite{bhunia2023person, zhang2023adding} to generate high-quality human images conditioned on 2D human poses. Such as ControlNet\cite{zhang2023adding}, which integrates conditional control into diffusion models, allowing precise manipulation of generated images based on pose inputs.

% TO BE DONE

% In our work, we propose a framework that utilizes the inherent generalizability of existing diffusion models to generate different poses of the same individual based on reference images and target pose information. By extracting features from these generated images, we aim to enhance the original features without any extra parameters. This approach focuses on improving feature quality during inference and can be seamlessly integrated into various ReID tasks, including standard, occluded, and cross-modality scenarios. It can be used with various SOTA ReID model.
\begin{figure*}
\centering
\includegraphics[width=0.85\textwidth]{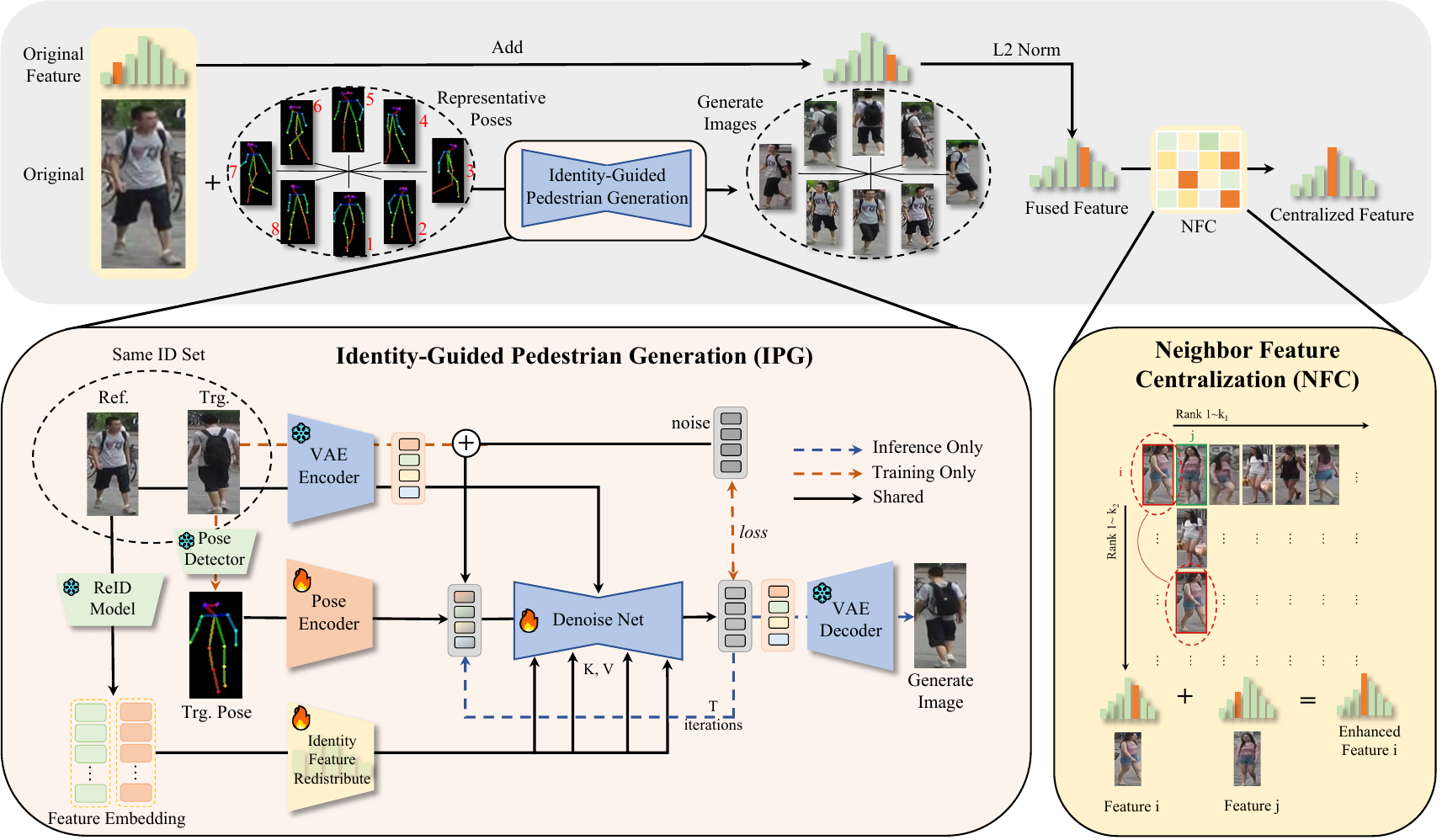}
\caption{Overview of the proposed Feature Centralization framework for the ReID task.}
\label{fig:main}
\end{figure*}

\section{Methods}
The main purpose of this paper is to \textbf{centralize features to their identity center} to enhance the identity representation of feature vectors extracted by ReID model, that is, reducing noise within the features while increasing the identity attributes to make them more representative of their identities. Therefore, to effectively and reasonably enhance the features, we need to understand the characteristics of the feature vectors obtained by ReID model.

\subsection{Feature Distribution Analysis} \label{Feature Distribution}

Currently, ReID models commonly use cross-entropy loss to impose ID-level constraints, and contrastive losses (such as triplet loss) to bring features of the same ID closer while pushing apart features of different IDs. Some models also utilize center loss to construct identity centers for dynamically constraining the IDs. These methods lead to one common result: feature aggregation. One can say that the current ReID task is essentially a feature aggregation task. The degree of feature density (e.g. t-SNE visualization) is also widely used to measure model performance. It is easy to deduce that the features of the same ID are centered around a "mean," approximately forming a normal distribution, as the distribution shown in Fig.\ref{fig:intro} which is visualized with one single feature dimension of the same ID.

It is evident that ReID features are normally distributed around the ‘identity center’. To theoretically prove that the feature vectors of each ID in current ReID tasks aggregation around a center or mean, we analyze several commonly used loss functions and their impact on the feature distribution in \textbf{Supplementary}.

For the same identity \( y_i = j \), we have a large number of samples \( \{\mathbf{x}_i\}_{i=1}^{N_j} \), where \( N_j \) is the number of samples for ID \( j \). These samples are passed through a ReID model \( f(\cdot) \), resulting in the corresponding feature vectors \( \{\mathbf{f}_i\}_{i=1}^{N_j} \):
\begin{equation}
\mathbf{f}_i = f(\mathbf{x}_i)
\end{equation}
where \( \mathbf{f}_i \in \mathbb{R}^d \) is the feature vector of sample \( \mathbf{x}_i \), and \( d \) is the dimensionality of the feature space.

For each feature dimension \( k \) of the feature vector \( \mathbf{f}_i \), the values \( \{\mathbf{f}_{i,k}\}_{i=1}^{N_j} \) obtained from different samples of the same ID \( j \) is independent and identically distributed random variables. Here, \( \mathbf{f}_{i,k} \) represents the \( k \)-th dimension of the feature vector for the \( i \)-th sample.

Since the input samples \( \{\mathbf{x}_i\}_{i=1}^{N_j} \) are independent, the values of \( \mathbf{f}_{i,k} \) are independent factors. According to the Central Limit Theorem (CLT), when the number of independent factors is large, the distribution of the values \( \{\mathbf{f}_{i,k}\}_{i=1}^{N_j} \) for any dimension \( k \) of the feature vector will approximate a normal distribution. Thus, for each feature dimension \( k \), we have:
\begin{equation}
\mathbf{f}_{i,k} \sim \mathcal{N}(\mu_k, \sigma_k^2)
\end{equation}
where \( \mu_k \) is the mean of the \( k \)-th feature dimension for ID f\( j \), and \( \sigma_k^2 \) is the variance of feature values in this dimension.

Since each dimension \( \mathbf{f}_{i,k} \) of the feature vector approximately follows a normal distribution across samples, the entire feature vector \( \mathbf{f}_i \) for ID \( j \) can be approximated by a multivariate normal distribution. This gives:
\begin{equation}
\mathbf{f}_i \sim \mathcal{N}(\boldsymbol{\mu}, \Sigma)
\end{equation}
where \( \boldsymbol{\mu} = (\mu_1, \mu_2, \dots, \mu_d)^\top \) is the mean vector of the feature dimensions, and \( \Sigma \) is the covariance matrix.

The theoretical analysis above suggests that under the optimization of the loss functions, the ReID model's feature vectors $\mathbf{x}_i$ aggregated around their identity centers $\mathbf{c}_{y_i}$ following a normal distribution. This is consistent with the feature aggregation observed in t-SNE visualizations.

\paragraph{Identity Density ($\text{ID}^2$) Metric}
Identity density is one aspect of measuring ReID effectiveness. However, there is currently no quantitative metric for this, and researchers commonly rely on visualization tools like t-SNE to demonstrate model performance. By using the concept above, we propose an Identity Density ($\text{ID}^2$) Metric which is detailed in \textbf{Supplementary}.
% That is,
% For a given identity \( y_i = i \), let \( \mathbf{z}_{i1}, \mathbf{z}_{i2}, \dots, \mathbf{z}_{in} \) represent the feature vectors extracted from different images of the same person. The mean feature vector \( \mathbf{\bar{z}}_i \), which serves as a more robust and representative feature for identity \( i \), can be computed as:
% \begin{equation}
% \mathbf{\bar{z}}_i = \frac{1}{n} \sum_{j=1}^{n} \mathbf{z}_{ij}
% \end{equation}

% In the ReID task, there are two main sets Gallery and Query, in the test of the dataset, is not allowed to utilize the a priori label information to assist in the recognition, which exists the information leakage, but in some particular scenarios, we may have multiple images of different poses or viewpoints of the searching target, it can be possible to overlay the features so that they are closer to the ID center to make it reduce the feature noise and be more ID representation.

% \subsection{Pose-based Pedestrians Generation}
\subsection{Feature Centralization via Identity-Guided Pedestrian Generation}

% In this section, we propose a novel method to centralize person features by generating images of the same individual (identity) with different poses using a Stable Diffusion model guided by identity feature. This approach leverages the conclusion above that features of the same identity follow a normal distribution, allowing for direct aggregation to centralize features to fit the identity center.

% Our purpose is to generate diverse images of a person under different poses and use these images to improve the feature representation of that person. 
\subsubsection{Feature Centralization}

Since features of the same identity follow a multivariate normal distribution, we can simply \textbf{aggregate features of the same identity to approximate the identity center}, as the visualization in Fig.\ref{fig:intro}. Thus, our main purpose becomes how to get more samples of the same identity to help centralize features.

\label{flip}
A straightforward approach is to perform horizontal flipping on the original images, and add features together. It is worth noting for reviewers that this is a very simple but effective trick. Therefore, it is necessary to check whether performance improvements are due to such tricks. In our experiments, to demonstrate the advancement of our approach, we did not use this trick. If used, it may be better. 

\subsubsection{Identity-Guided Diffusion Process}
To get more samples from the same identity, we propose a novel Identity-Guided Pedestrian Generation (IPG) paradigm, generating images of the same identity with different poses using a Stable Diffusion model guided by identity feature to centralize each sample's features. 
 
Followed by Stable Diffusion \cite{rombach2022high}, which is developed from latent diffusion model (LDM). We use the reference UNet to inject the reference image features into the diffusion process with a constant timestep \( t = 0 \), and the denoising UNet $\boldsymbol{\epsilon}_{\theta}$ to generate the target image latent \(\mathbf{z_0} \) by denoising the noisy latent \( \mathbf{z_T} = \boldsymbol{\epsilon}\):
\begin{equation}
\mathbf{z}_{t-1}=\boldsymbol{\epsilon}_{\theta}(\mathbf{z}_t, t, \mathbf{E}_{\text{pose}}, \mathbf{H}), t \in [0,T]
\end{equation}
where \( \mathbf{H} \) is the identity feature information to guide model keep person identity. \( \mathbf{E}_{\text{pose}} \) is the pose features.

At each timestep \( t \), the pose feature \( \mathbf{E}_{\text{pose}} \) and conditioning embedding \( \mathbf{H} \) guide the denoising process.
\\
\textbf{Identity Feature Redistribute (IFR)}
The Identity Feature Redistribute (IFR) module aims to utilize identity features from input images, removing noise to better guide the generative model. It converts high-dimensional identity features into meaningful low-dimensional feature blocks, enhancing the model’s feature utilization efficiency.

Given the input sample $\mathbf{x} \in \mathbb{R}^{C}$ by a ReID model \( f(\cdot) \), with IFR, we can obtain a re-distributed robust feature $\mathbf{H}\in \mathbb{R}^{N \times D})$ :
\begin{gather}
\mathbf{H} = \text{IFR}(f(\mathbf{x})) = \text{LN}(\text{Linear}(\mathbf{f}))
\end{gather}

For this more robust feature identity feature, it is used as the K, V of the model's attention module to guide the model's attention to the identity feature.
\\
\textbf{Pose Encoder}
The Pose Encoder is to extract high-dimensional pose embeddings $\mathbf{E}_{\text{pose}}$ from input poses. It has 4 blocks with 16,32,64,128 channels. Each block applies a normal \(3 \times 3\) Conv, a \(3 \times 3\) Conv with stride 2 to reduce spatial dimensions, and followed by a SiLU activate function. Subsequently, the pose features are added to the noise latent before into the denoising UNet, follows \cite{hu2024animate}.
\\
\textbf{Training Strategy}
For each identity $i$, we randomly select one image from \( S_i^{\text{ref}} \) as the reference image and one image from \( S_i^{\text{trg}} \) as the target image for training.

Let \( \mathbf{x}_{i,j}^{\text{ref}} \in S_i^{\text{ref}} \) denote the reference image (i.e. $j_{th}$ image of the $i_{th}$ ID) and \( \mathbf{x}_{i,j}^{\text{trg}} \in S_i^{\text{trg}} \) denote the target image. Model is trained using the mean squared error (MSE) loss between the predicted noise and the true noise.
\begin{equation}
\mathcal{L} = \mathbb{E}_{\mathbf{z}, t, \boldsymbol{\epsilon}} \left[ \left\| \boldsymbol{\epsilon} - \boldsymbol{\epsilon})\theta(\mathbf{z}_t, t, \mathbf{E}_{\text{pose}}, \mathbf{H}) \right\|^2 \right]
\end{equation}
where $\mathbf{z}=\text{VAE}(\mathbf{x}^{\text{trg}}_{i,j})+\boldsymbol{\epsilon}$ is a latent obtained by a pre-trained VAE encoder\cite{kingma2013auto}, $\mathbf{E}_{\text{pose}}$ is the pose feature of $\mathbf{x}^{\text{trg}}_{i,j})$, $\mathbf{H}$ is the re-distributed identity feature of $(\mathbf{x}^{\text{ref}}_{i,j}))$.

The model is trained to learn the mapping from the reference image \( \mathbf{x}_{\text{ref}} \) to the target image \( \mathbf{x}_{\text{trg}} \), with the goal of generating realistic variations in pose while preserving identity feature. This random selection ensures diversity during training, as different combinations of reference and target images are used in each training iteration, enhancing the model’s ability to generalize across various poses and viewpoints.

\subsubsection{Selection of Representative Pose} \label{standard pose}

In ReID tasks, features extracted from different poses of the same identity can vary significantly. Some specific poses tend to be more representative of that identity. As the conclusion of feature distribution in section\ref{Feature Distribution}, we calculate the identity center for IDs with all of its samples in datasets, and select the image whose feature is the closest to the center. The pose of this image is regarded as the representative pose. By randomly selecting 8 representative poses with different directions, we generate images that are more representative of the person’s identity.

That is, given a set of feature vectors \( \mathbf{F}_{\text{all}} = \{ \mathbf{f}_1, \mathbf{f}_2, \dots, \mathbf{f}_N \} \) for a particular identity:
\begin{equation}
\mathbf{f}_{\text{mean}} = \frac{1}{N} \sum_{i=1}^{N} \mathbf{f}_i
\end{equation}
\begin{equation}
\text{pose} = \arg \min_{i}d(\mathbf{f}_{\text{mean}}, \mathbf{f}_i) 
\end{equation}

\subsubsection{Feature Centralization Enhancement}
Once we got generated images with different poses, we generate new images \( \hat{\mathbf{x}} \) for each of these poses. The features extracted from these generated images are then aggregated with the original reference feature to enhance the overall representation. The centralized feature \(\tilde{\mathbf{f}}\) is computed as:
\begin{equation}
\tilde{\mathbf{f}}  = \|\mathbf{f} + \frac{\eta}{M} \sum_{i=1}^{M} \mathbf{f}_{i}\|_2
\end{equation}
where \( \mathbf{f}\) is the feature of the original reference image, and \( \mathbf{f}_{i} \) are the features extracted from the \( M \) generated images. The coefficient \( \eta \) is introduced to adjust based on the quality of generated images. According to the theory of Section\ref{Feature Distribution}, low-quality generated images, as long as they contain corresponding identity information, can also be applied with feature enhancement, and use \( \eta \) to regulate the enhancement effect of generated information. As discussed in \cite{zheng2017unlabeled}, even if quality is poor, it still contains ID information.

% This strategy ensures that generated images contribute to feature aggregation without introducing significant discrepancies, particularly for models that are sensitive to part details. By focusing on the most similar poses and aggregating their features, we can create a more robust and representative feature for each sample.
\subsection{Neighbor Feature Centralization (NFC)}

Moreover, we proposed a \textbf{Neighbor Feature Centralization (NFC)} algorithm to reduce noise in individual features and improve their identity discriminability in unlabeled scenarios. The core idea of the algorithm is to utilize mutual nearest-neighbor features for aggregation. 
% It still focuses on the main purpose of this paper: \textbf{utilize as much as the poses have for ID Representation}. 
\begin{algorithm}[H]
\caption{Neighbor Feature Centralization (NFC)}
\label{alg:feature_enhancement}
\begin{algorithmic}[1] 
\Require Feature vectors \(\{\mathbf{z}_i\}_{i=1}^N\), parameters \(k_1\), \(k_2\)
\Ensure Centralized feature vectors \(\{\mathbf{z}_i^{\text{centralized}}\}_{i=1}^N\)

\State Compute pairwise distance matrix \(\mathbf{D} = [d_{ij}]\)
\For{\(i = 1\) to \(N\)}
    \State Set \(d_{ii} = C\) \Comment{Avoid self-matching}
\EndFor
\For{\(i = 1\) to \(N\)}
    \State \(\mathcal{N}_{i} \gets\) indices of top \(k_1\) neighbors of \(\mathbf{z}_i\)
\EndFor
\For{\(i = 1\) to \(N\)}
    \State \(\mathcal{M}_{i} \gets \emptyset\)
    \For{each \(j \in \mathcal{N}_{i}\)}
        \State \(\mathcal{N}_{j}^{k_2} \gets\) indices of top \(k_2\) neighbors of \(\mathbf{z}_j\)
        \If{\(i \in \mathcal{N}_{j}^{k_2}\)}
            \State \(\mathcal{M}_{i} \gets \mathcal{M}_{i} \cup \{ j \}\)
        \EndIf
    \EndFor
\EndFor
\For{\(i = 1\) to \(N\)}
    \State \(\mathbf{z}_i^{\text{centralized}} \gets \mathbf{z}_i + \sum_{j \in \mathcal{M}_{i}} \mathbf{z}_j\)
\EndFor
\end{algorithmic}
\end{algorithm}
By enhancing each feature with its potential neighbors, it could effectively approximate features of the same identity without explicit labels, and ensure that only features have high similarity relationships contribute to the enhancement.

% \section{}

\begin{table*}[ht]
\centering
\small
\renewcommand{\arraystretch}{0.95}
\renewcommand\tabcolsep{4pt}
\begin{tabular}{cc|c|c|c|c|ccc}
\hline 
\multicolumn{2}{c|}{Dataset} & Model & Venue & Base & Method & mAP$\uparrow$ & Rank-1$\uparrow$ & $\text{ID}^2$$\downarrow$ \\ \hline
% \multicolumn{2}{c|}{} & & & & ori & 3.34 & 11.4 & 0.5135 \\ \cline{6-9} 
% \multicolumn{2}{c|}{} & \multirow{-2}{*}{TransReID\cite{he2021transreid}(w/o training)} & & & +ours & 52.81\scriptsize{(+49.47)} & 78.92\scriptsize{(+67.52)} & 0.2158 \\ \cline{3-3} \cline{6-9} 
\multicolumn{2}{c|}{} & & & & official & 79.88 & 91.48 & 0.2193 \\ \cline{6-9} 
\multicolumn{2}{c|}{} & \multirow{-2}{*}{TransReID\cite{he2021transreid}(w/o camid)} & & & +ours & 90.39\scriptsize{(+10.51)} & 94.74\scriptsize{(+3.26)} & 0.1357 \\ \cline{3-3} \cline{6-9} 
\multicolumn{2}{c|}{} & & & & official & 89 & 95.1 & 0.2759 \\ \cline{6-9} 
\multicolumn{2}{c|}{} & \multirow{-2}{*}{TransReID\cite{he2021transreid}(w/ camid)} & \multirow{-4}{*}{ICCV21} & \multirow{-4}{*}{ViT} & +ours & 93.01\scriptsize{(+4.01)} & 95.52\scriptsize{(+0.42)} & 0.1967 \\ \cline{3-9} 
\multicolumn{2}{c|}{} & & & & official & 89.7 & 95.4 & 0.0993 \\ \cline{6-9} 
\multicolumn{2}{c|}{} & & & \multirow{-2}{*}{ViT} & +ours & 94\scriptsize{(+4.3)} & 96.4\scriptsize{(+1.0)} & 0.0624 \\ \cline{5-9} 
\multicolumn{2}{c|}{} & & & & official & 89.8 & 95.7 & 0.0877 \\ \cline{6-9} 
\multicolumn{2}{c|}{\multirow{-8}{*}{Market1501}} & \multirow{-4}{*}{CLIP-ReID\cite{li2023clip}} & \multirow{-4}{*}{AAAI23} & \multirow{-2}{*}{CNN} & \cellcolor{gray!20}+ours & \cellcolor{gray!20}94.9\scriptsize{(+5.1)} & \cellcolor{gray!20}97.3\scriptsize{(+1.6)} & \cellcolor{gray!20}0.053 \\ \hline
\multicolumn{2}{c|}{} & & & & official & 79.05 & 85.4 & 0.3124 \\ \cline{6-9} 
\multicolumn{2}{c|}{} & \multirow{-2}{*}{KPR\cite{somers2025keypoint}} & \multirow{-2}{*}{ECCV24} & \multirow{-2}{*}{ViT} & \cellcolor{gray!20}+ours & \cellcolor{gray!20}89.34\scriptsize{(+10.29)} & \cellcolor{gray!20}91\scriptsize{(+5.6)} & \cellcolor{gray!20}0.1434 \\ \cline{3-9} 
\multicolumn{2}{c|}{} & & & & official & 70.41 & 77.2 & 0.377 \\ \cline{6-9} 
\multicolumn{2}{c|}{\multirow{-4}{*}{Occluded-ReID}} & \multirow{-2}{*}{BPBReID\cite{somers2023body}} & \multirow{-2}{*}{WACV23} & \multirow{-2}{*}{ViT} & +ours & 86.05\scriptsize{(+15.64)} & 89.1\scriptsize{(+11.9)} & 0.1504 \\ \hline
\multicolumn{1}{c|}{} & & & & & official & 71.81 & 75.29 & 0.4817 \\ \cline{6-9} 
\multicolumn{1}{c|}{} & \multirow{-2}{*}{All} & & & & \cellcolor{gray!20}+ours & \cellcolor{gray!20}76.44\scriptsize{(+4.63)} & \cellcolor{gray!20}79.33\scriptsize{(+4.04)} & \cellcolor{gray!20}0.4072 \\ \cline{2-2} \cline{6-9} 
\multicolumn{1}{c|}{} & & & & & official & 84.6 & 81.59 & 0.4424 \\ \cline{6-9} 
\multicolumn{1}{c|}{} & \multirow{-2}{*}{Indoor} & \multirow{-4}{*}{SAAI\cite{fang2023visible}} & \multirow{-4}{*}{ICCV23} & \multirow{-4}{*}{CNN} & \cellcolor{gray!20}+ours & \cellcolor{gray!20}86.83\scriptsize{(+2.23)} & \cellcolor{gray!20}84.2\scriptsize{(+2.61)} & \cellcolor{gray!20}0.3694 \\ \cline{2-9} 
\multicolumn{1}{c|}{} & & & & & official & 66.13 & 67.7 & 0.4308 \\ \cline{6-9} 
\multicolumn{1}{c|}{} & \multirow{-2}{*}{All} & & & & +ours & 75.43\scriptsize{(+9.3)} & 74.81\scriptsize{(+7.11)} & 0.3133 \\ \cline{2-2} \cline{6-9} 
\multicolumn{1}{c|}{} & & & & & official & 77.81 & 72.95 & 0.4046 \\ \cline{6-9} 
\multicolumn{1}{c|}{\multirow{-8}{*}{SYSU-MM01}} & \multirow{-2}{*}{Indoor} & \multirow{-4}{*}{PMT\cite{lu2023learning}} & \multirow{-4}{*}{AAAI23} & \multirow{-4}{*}{ViT} & +ours & 84.29\scriptsize{(+6.48)} & 80.29\scriptsize{(+7.34)} & 0.2995 \\ \hline
\end{tabular}

\caption{Improvements with our method on different SOTA models with both ViT and CNN backbone on Market1501, SYSU-MM01, and Occluded-ReID datasets. The data under \colorbox{gray!20}{grey} is the new SOTA with our methods of that dataset.}
\label{tab:sota}
\end{table*}

\begin{table*}[ht]
\centering
\small
\renewcommand{\arraystretch}{1.0}
\renewcommand\tabcolsep{4pt}
\begin{minipage}{0.32\textwidth}
\centering
\begin{tabular}{c|ccc}
\hline
\textbf{Methods} & mAP$\uparrow$ & Rank-1$\uparrow$ & $\text{ID}^2$$\downarrow$ \\ \hline
Base & 79.88 & 91.48 & 0.2193 \\
+NFC & 83.92 & 91.83 & 0.1824 \\
+IPG & 88.02 & 94.77 & 0.1553 \\
\rowcolor{gray!20}
+NFC+IPG & 90.39 & 94.74 & 0.1357 \\
\hline
\end{tabular}
\caption{Ablation study on effects of feature centralization through Identity-Guided Pedestrian Generation (IPG) and Neighbor Feature Centralization (NFC).}
\label{tab:ablation_fe_pg}
\end{minipage}
\hfill
\begin{minipage}{0.32\textwidth}
\centering
\begin{tabular}{cc|cc}
\hline
$\text{Gallery}^\text{NFC}$ & $\text{Query}^\text{NFC}$  & mAP$\uparrow$ & Rank-1$\uparrow$ \\ \hline
 \ding{55} & \ding{55}   & 79.88&	91.48  \\
 \ding{51} & \ding{55}    & 81.70 & 92.04  \\
\ding{55} & \ding{51}    & 82.76 & 91.69  \\
\rowcolor{gray!20}
\ding{51} & \ding{51}    & 83.92 & 91.83 \\ 
\hline
\end{tabular}
\caption{Ablation study of Neighbor Feature Centralization (NFC) Algorithm on Market1501 dataset. We test on the gallery and query set respectively.}
\label{tab:as_fe}
\end{minipage}
\hfill
\begin{minipage}{0.32\textwidth}
\centering
\begin{tabular}{cc|cc}
\hline
% \multicolumn{2}{c|}{\textbf{Methods}}& \multicolumn{4}{c}{Metrics}  \\ \hline
$\text{Gallery}^\text{IPG}$ & $\text{Query}^\text{IPG}$  & mAP$\uparrow$ & Rank-1$\uparrow$ \\ \hline
\ding{55} & \ding{55}   & 79.88&	91.48  \\
% \ding{51} & \ding{55}    &  84.06 &91.48 \\
% \ding{55} & \ding{51}    & 81.01 &91.03  \\
% \rowcolor{gray!20}
% \ding{51} & \ding{51}    & 87.38& 94.63 \\ 
 \ding{51} & \ding{55}    & 84.65&	92.07  \\
\ding{55} & \ding{51}    & 82.18	&92.40  \\
\rowcolor{gray!20}
\ding{51} & \ding{51}    &88.02&	94.77 \\ 
\hline
\end{tabular}
\caption{Ablation study of Feature ID-Centralizing with Pedestrian Generation (IPG) on Market1501. We test on gallery and query set respectively.}
\label{tab:as_gen}
\end{minipage}
\end{table*}

\begin{table}
\small
    \centering
    \renewcommand{\arraystretch}{1}
    \renewcommand\tabcolsep{6pt}
    \begin{tabular}{c|ccccc}
        \hline
        \textbf{Method} & mAP$\uparrow$ & R1$\uparrow$ & R5$\uparrow$ & R10$\uparrow$ & $\text{ID}^2$$\downarrow$ \\ \hline
        w/o training & 3.34 & 11.4 & 21.88 & 28 & 0.5135 \\
        \rowcolor{gray!20}
        +IPG & 52.81 & 78.92 & 91.21 & 94.27 & 0.2158 \\
        \rowcolor{gray!20}
        +IPG+NFC & 57.27 & 82.39 & 90.17 & 92.81 & 0.1890 \\ \hline
    \end{tabular}
    \caption{ReID performance on Market1501 with only ImageNet pre-trained weights without ReID training. The distribution visualized in Fig.\ref{fig:noise_tsne}.}
    \label{tab:notraining}
\end{table}
\begin{figure}[H]
\centering
\includegraphics[width=0.95\linewidth]{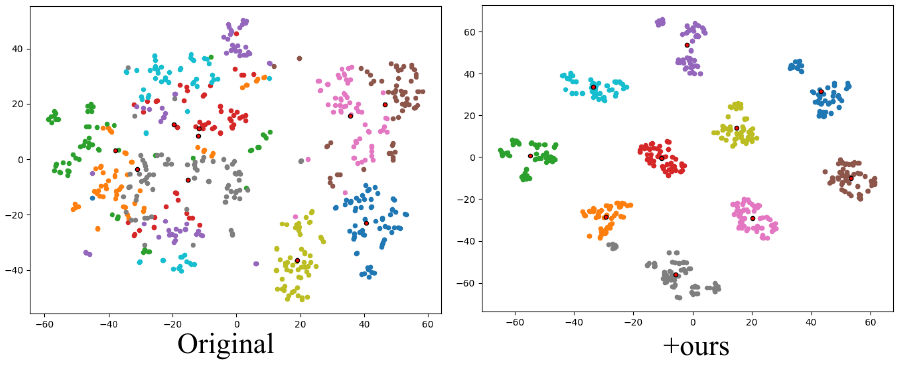}
\caption{t-SNE visualization of 10 IDs feature distribution with and without our method on ImageNet pre-trained weights.}
\label{fig:noise_tsne}
\end{figure}

\section{Experiments}
\subsection{Implementation Details}

\textbf{Data Cleaning.}
Training an effective generative model requires high-quality data support. In current ReID (Person Re-Identification) datasets, there are many low-quality images, and removing them can help reduce interference to the model. In our experiments, we found two main issues that need to be addressed:\textbf{Extremely Low-quality Images}: The dataset contains images with such low resolution that even the human eye cannot recognize them as a "person". \textbf{Pose Estimation Failures}: The pose estimation model inevitably fails to detect pedestrian poses in some images.

Utilizing the feature distribution mentioned in Section\ref{Feature Distribution}, we can solve the issues and get the reference image set $S_i^{\text{ref}}$ and target image set $S^{\text{trg}}_i$ of identity $i$.
The cleansing process is detailed in \textbf{Supplementary}. 
\\
\textbf{Identity-guided Pedestrian Generation model details}
For the reference UNet and denoising UNet, we use the pretrained weights of Stable Diffusion v1.5\cite{rombach2022high}. The VAE encoder and decoder initialized with official weights\cite{kingma2013auto} and froze. For the ReID model, we use pre-trained TransReID\cite{he2021transreid} without cameras on Market1501 and freeze. 

The training data collected from training set of Market1501\cite{zheng2015scalable},  Mars\cite{zheng2016mars}, MSMT17\cite{wei2018person} and SYSU-MM01\cite{wu2017rgb}, with a total of 1946 different IDs. The model was trained for 80,000 iterations on one L20(48G) GPU with batch size of 4, which costs about 20 hours, and optimized by Adam\cite{kingma2014adam} with a learning rate of 1e-5 and weight decay of 0.01. All images are resized to 512×256. We applied random flip and random erasing\cite{zhong2020random} data augmentation only on reference images.

According to Section\ref{standard pose}, we selected 8 poses on the market1501 dataset as the representative poses. Each image in test set generates 8 images with these representative poses. The generation uses DDIM with 20 steps, classifier-free guidance with a scale of 3.5, and generator seed of 42.
\\
\textbf{ReID test settings}
Test models are loaded with official open-source pre-trained models for testing. In addition, considering the generated images do not have camera IDs, so for feature consistency, we test without camera IDs (e.g. TransReID). To validate the effectiveness of our proposed method, image flip (Section\ref{flip}) trick is \textbf{NOT} applied in our experiments. On Market1501, set \(\eta=2\), and \(k_1=k_2=2\). On Occluded REID, set \(\eta=1\), and \(k_1=k_2=1\). On SYSU-MM01, set \(\eta=1/4\), and \(k_1=k_2=2\). The parameters analysis detailed in \textbf{Supplementary}.

\subsection{Improvements on State-of-the-art Methods}
To verify the exceptional feature enhancement performance of our framework, we selected state-of-the-art models of three ReID tasks, divided into CNN and ViT-based models, to demonstrate that our theory can apply to any model and various ReID tasks. As shown in Table.\ref{tab:sota}, we achieved excellent enhancements on different models.

It is worth mentioning that we help models achieve new \textbf{SOTAs} without re-rank on 3 benchmarks:
\begin{itemize}
    \item[$\bullet$] \textbf{CNN-base CLIP-ReID on market1501: }
    
    \setlength{\parindent}{1em} mAP 94.9\%, Rank-1 97.3\%. 
    \item[$\bullet$] \textbf{KPR on Occluded-ReID: }

    \setlength{\parindent}{1em} mAP 89.34\%, Rank-1 91\%
    
    \item[$\bullet$] \textbf{SAAI on SYSU-MM01: }

    \setlength{\parindent}{1em} All-search mode: mAP 76.44\%, Rank-1 79.33\%
    
    \setlength{\parindent}{1em} Indoor-search mode: mAP 86.83\%, Rank-1 84.2\%
\end{itemize}

\subsection{ReID without Training}
TransReID loads a ViT pre-trained model on ImageNet for training on the ReID task. The pre-training method is based on contrastive learning strategies. According to the description in Section\ref{Feature Distribution}, training with contrastive loss helps to cluster features of same label samples. Images generated by our Pedestrian Generation model exhibit identity consistency, meaning they possess the attributes of the same label. Therefore, even if the features of an individual sample lack the pedestrian matching capability, as shown in the first row of Tab.\ref{tab:notraining}, its mAP and Rank-1 are only 3.34\% and 11.4\%. However, with our method, it improved 53.93\%/70.99\% to 57.27\%/82.39\%. Additionally, we visualized of 10 IDs' feature distributions using t-SNE as shown in Fig.\ref{fig:noise_tsne}.

\begin{figure}
\centering
\includegraphics[width=0.95\linewidth]{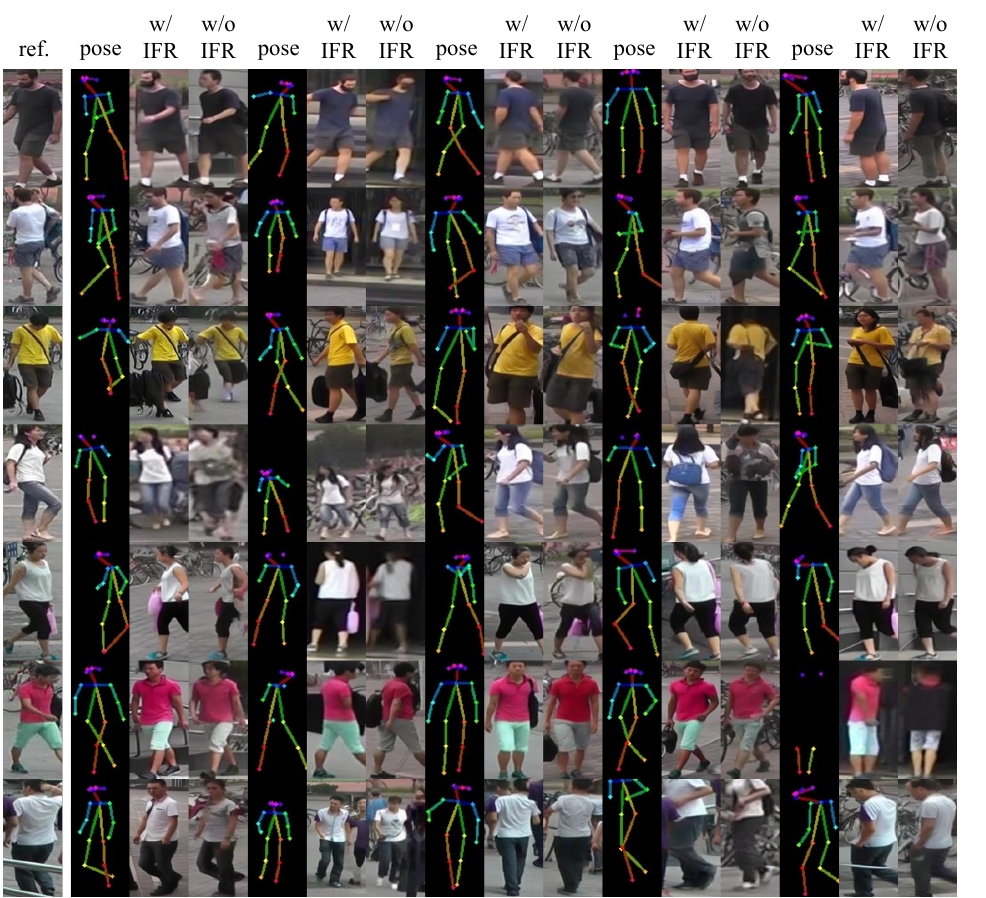}
\caption{The effects with and without the IFR module were visualized with five different poses randomly selected for each reference picture.}
\label{fig:IFR}
\end{figure}

\begin{table}
\small
    \centering
    \renewcommand{\arraystretch}{1}
    \renewcommand\tabcolsep{2pt}
    \begin{tabular}{c|ccc|ccc|ccc}
        \hline
        \multirow{2}{*}{\textbf{N}} & \multicolumn{3}{c|}{$\text{Gallery}^\text{IPG}$} & \multicolumn{3}{c|}{$\text{Query}^\text{IPG}$} & \multicolumn{3}{c}{$\text{Gallery}^\text{IPG}$+$\text{Query}^\text{IPG}$} \\ \cline{2-10}
          & mAP$\uparrow$ & R1$\uparrow$ & $\text{ID}^2$$\downarrow$ & mAP$\uparrow$ & R1$\uparrow$ & $\text{ID}^2$$\downarrow$ & mAP$\uparrow$ & R1$\uparrow$ & $\text{ID}^2$$\downarrow$ \\ \hline
        0 & 79.88 & 91.48 & 0.2313 & 79.88 & 91.48 & 0.1623 & 79.88 & 91.48 & 0.2193 \\
        1 & 80.96 & 90.97 & 0.2087 & 79.99 & 90.83 & 0.1363 & 82.13 & 92.01 & 0.1961 \\
        2 & 82.86 & 91.45 & 0.1904 & 81.17 & 92.04 & 0.1156 & 85.27 & 93.71 & 0.1773 \\
        3 & 83.42 & 91.75 & 0.1837 & 81.55 & 92.25 & 0.1077 & 86.16 & 94.21 & 0.1704 \\
        4 & 83.81 & 92.1  & 0.1795 & 81.76 & 92.34 & 0.1027 & 86.75 & 94.24 & 0.1661 \\
        5 & 84.07 & 91.83 & 0.1758 & 81.85 & 92.07 & 0.0984 & 87.23 & 94.71 & 0.1623 \\
        6 & 84.3  & 91.98 & 0.1730 & 81.96 & 92.52 & 0.0950 & 87.52 & 94.63 & 0.1593 \\
        7 & 84.49 & 91.86 & 0.1707 & 82.03 & 92.37 & 0.0922 & 87.76 & 94.60 & 0.1570 \\
        \rowcolor{gray!30}
        8 & 84.65 & 92.07 & 0.1691 & 82.18 & 92.40 & 0.0902 & 88.02 & 94.77 & 0.1553 \\ \hline
    \end{tabular}
    \caption{Performance Comparison by adding numbers of generated images for each image on gallery, query, and both}
    \label{tab:num}
\end{table}

\subsection{Abilation Study}
\textbf{Impact of the NFC and IPG.}
We conducted comprehensive ablation experiments on Neighbor Feature Centralization (NFC) and Feature ID-Centralizing through Identity-Guided Generation (IPG) methods. As shown in Table\ref{tab:ablation_fe_pg},\ref{tab:as_fe},\ref{tab:as_gen}, which show great improvements.
\begin{figure*}
\centering
\includegraphics[width=\linewidth]{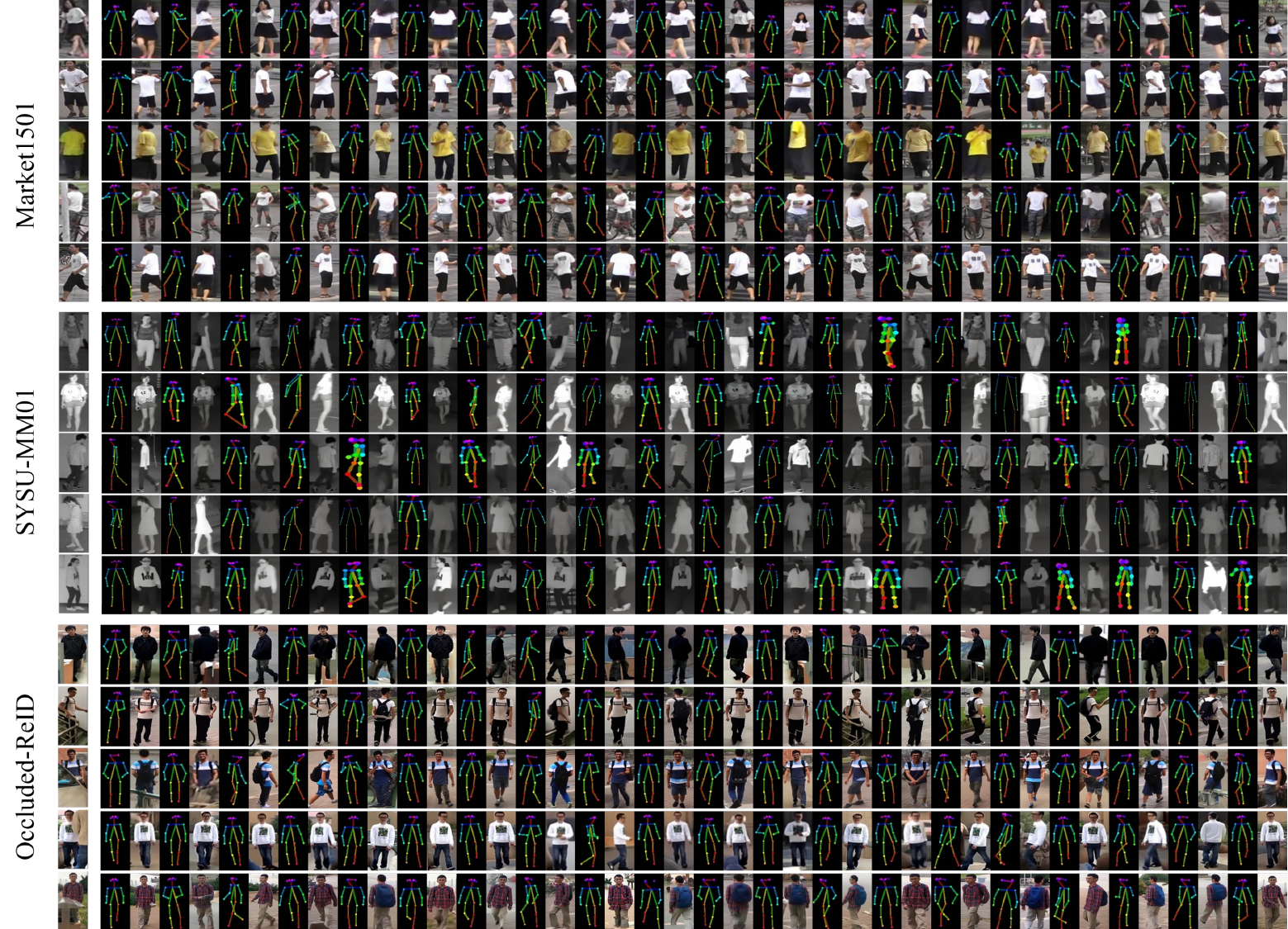}
\caption{Images generated with random poses. More randomly generated images can be found in Supplementary.}
\label{fig:randompose}
\end{figure*}
\\
\textbf{Effect of Feature Re-Distribute Module.}
We randomly selected 7 images from the Market1501 dataset, choosing 5 different poses for each image. We visualized the results both with and without Identity Feature Redistribute (IFR) Module. As shown in Fig.\ref{fig:IFR}, the impact of ID features on the generated outcomes is evident.
\\
\textbf{Effect of numbers of generated images.}
We randomly selected different numbers of images generated from the 8 representative poses to verify the effect of feature enhancement. As shown in Tab.\ref{tab:num}, the experimental results align with the theory mentioned in Section\ref{Feature Distribution}: the more features of the same ID that are aggregated, the more the adjustment noise extracted from individual images is reduced, enhancing the ID representation capability and resulting in improved matching performance.

\subsection{Visualizations}

\textbf{Different people with the same poses across datasets.}
We are working on Market1501, SYSU-MM01, and
Occluded-ReID datasets to visualize the 8 representative poses with only one model and results are shown in Fig.\ref{fig:intro}.
\\
\textbf{Random people with random poses.} To demonstrate the advancement of our model, as shown in Fig.\ref{fig:randompose}, we randomly chose samples from the whole dataset, and each sample randomly chose poses, including some problematic poses, fully demonstrating the diversity of model. \textbf{More examples on three datasets are visualized in }\textbf{Supplementary}.

% \subsection{Analysis on quality coefficient $\eta$ of Generation Model}

% Fig.\ref{fig:eta} illustrates the effect of adjusting the coefficient $\eta$ on the performance of the ReID model. To evaluate this impact, we gradually increased the value of $\eta$ and observed the changes in model performance on the mAP and Rank-1 metrics. 

% As the value of $\eta$ increases, the performance of the ReID model improves, reaching an optimal point. At $\eta = 2$, both mAP and Rank-1 achieve their maximum values of 88.02\% and 94.77\%, respectively. However, further increasing $\eta$ beyond this point leads to a slight decline in performance. It is easy to find that using generated images to centralize features is effective. However, considering the quality of the generated image, direct adding, although also effective, may not always achieve the best results. Therefore adjusting $\eta$ according to the generation quality of the model in this dataset can better centralize the features.
% \begin{figure}
% \centering
% \includegraphics[width=0.9\linewidth]{figs/pdf/eta.pdf}
% \caption{Impact of the quality coefficient \( \eta \) with TransReID on Market1501. The dark color lines are the baseline.}
% \label{fig:eta}
% \end{figure}
\section{Conclusion}
In this paper, we proposed a training-free person re-identification framework that fully leverages the natural clustering behavior of features around the identity center during the training process. By introducing the concept of feature centralization, we effectively reduced noise in individual samples and enhanced identity representation without model training. Our approach includes an identity-guided pedestrian generation paradigm, capable of producing high-quality, multi-pose images with consistent identity features, even under challenging conditions such as visible, infrared, and occlusion scenarios. The neighbor feature centralization algorithm also preserves feature's original distribution while mining potential positive samples. It can also be flexibly integrated with existing re-ranking methods.

% \section*{Acknowledgments}
% This work was supported by the National Natural Science Foundation of China (No. 62376016).
% \input{sec/X_suppl_arxiv}
{
    \small
    \bibliographystyle{ieeenat_fullname}
    \bibliography{main}
}

% \setcounter{section}{0}
\clearpage
\setcounter{page}{1}
\maketitlesupplementary

\section{Methods Supplementary}
\subsection{Aggregation role of ReID loss functions}
Currently, ReID models commonly use cross-entropy loss to impose ID-level constraints, and contrastive losses (such as triplet loss) to bring features of the same ID closer while pushing apart features of different IDs. Some models also utilize center loss to construct identity centers for dynamically constraining the IDs. These methods lead to one common result: feature aggregation. From the perspective of the gradient of the loss functions, we could prove that the feature vectors of each ID in current ReID tasks naturally aggregate around a center or mean in the followings.
\paragraph{Cross-Entropy Loss} 
is often used in classification tasks, optimizing the model by maximizing the probability of the correct class. Given $N$ samples, each with a feature vector $\mathbf{z}_i \in \mathbb{R}^d$, and its corresponding class label $y_i \in \{1, 2, \dots, C\}$, the cross-entropy loss is defined as:
\begin{equation}
\mathcal{L}_{\text{CE}} = -\frac{1}{N} \sum_{i=1}^{N} \log \frac{\exp(\mathbf{w}_{y_i}^\top \mathbf{z}_i + b_{y_i})}{\sum_{j=1}^{C} \exp(\mathbf{w}_j^\top \mathbf{z}_i + b_j)}
\end{equation}
where $\mathbf{w}_j$ and $b_j$ are the weight vector and bias for class $j$, respectively.

For simplicity, assume that the final layer is a linear classifier without bias, i.e., $b_j = 0$. When the loss is minimized, the optimization objective is to maximize the score $\mathbf{w}_{y_i}^\top \mathbf{z}_i$ of the correct class while minimizing the scores $\mathbf{w}_j^\top \mathbf{z}_i$ of other classes ($j \neq y_i$).

By gradient descent optimization, we can obtain:
\begin{equation}
\frac{\partial \mathcal{L}_{\text{CE}}}{\partial \mathbf{z}_i} = 1/N\left(p_{y_i} - 1\right) \mathbf{w}_{y_i} + 1/N\sum_{j \neq y_i} p_{ij} \mathbf{w}_j
\end{equation}
where $p_{ij} = \frac{\exp(\mathbf{w}_j^\top \mathbf{z}_i)}{\sum_{k=1}^{C} \exp(\mathbf{w}_k^\top \mathbf{z}_i)}$.

With the loss function converges, $p_{y_i}\rightarrow1$ and $p_{ij}\rightarrow0 (j \neq y_i)$. The feature $\mathbf{z}_i$ is optimized to be near a linear combination of the class weight vectors $\mathbf{w}_{y_i}$. This indicates that features of the same class will tend toward a common direction, thus achieving feature aggregation.

\paragraph{Contrastive loss} (Triplet Loss as example) optimizes the feature space by bringing samples of the same class closer and pushing samples of different classes further apart. A triplet $(\mathbf{z}_a, \mathbf{z}_p, \mathbf{z}_n)$ is defined, where $\mathbf{z}_a$ is the anchor, $\mathbf{z}_p$ is the positive sample (same class), and $\mathbf{z}_n$ is the negative sample (different class). The triplet loss is defined as:
\begin{equation}
\mathcal{L}_{\text{Triplet}} = \max \left( \| \mathbf{z}_a - \mathbf{z}_p \|_2^2 - \| \mathbf{z}_a - \mathbf{z}_n \|_2^2 + \alpha, 0 \right)
\end{equation}
where $\alpha$ is the margin parameter.

To minimize the loss, the optimization objective is:
\begin{equation}
\| \mathbf{z}_a - \mathbf{z}_p \|_2^2 + \alpha < \| \mathbf{z}_a - \mathbf{z}_n \|_2^2
\end{equation}
\begin{align}
\frac{\partial \mathcal{L}_{\text{Triplet}}}{\partial \mathbf{z}_a} &= 2 (\mathbf{z}_n - \mathbf{z}_p), \\
\frac{\partial \mathcal{L}_{\text{Triplet}}}{\partial \mathbf{z}_p} &= 2 (\mathbf{z}_p - \mathbf{z}_a), \\
\frac{\partial \mathcal{L}_{\text{Triplet}}}{\partial \mathbf{z}_n} &= 2 (\mathbf{z}_a - \mathbf{z}_n).
\end{align}

By minimizing triplet loss, the feature $\mathbf{z}_p$ is pulled closer to $\mathbf{z}_a$, while $\mathbf{z}_n$ is pushed away. Through this mechanism, Triplet Loss encourages features of the same class to aggregate together while features of different classes are separated from each other.

\paragraph{Center loss} further enhances feature aggregation by introducing a feature center for each class. For each class $j$, there is a feature center $\mathbf{c}_j$, and the center loss is defined as:
\begin{equation}
\mathcal{L}_{\text{Center}} = \frac{1}{2} \sum_{i=1}^{N} \| \mathbf{z}_i - \mathbf{c}_{y_i} \|_2^2
\end{equation}

The goal of minimizing center loss is to make each sample's feature vector $\mathbf{z}_i$ as close as possible to its corresponding class center $\mathbf{c}_{y_i}$. Through gradient descent, we obtain:
\begin{align}
\frac{\partial \mathcal{L}_{\text{Center}}}{\partial \mathbf{z}_i} &= \mathbf{z}_i - \mathbf{c}_{y_i} \\
\frac{\partial \mathcal{L}_{\text{Center}}}{\partial \mathbf{c}_j} &= \begin{cases}
\mathbf{c}_j - \mathbf{z}_i & \text{if } y_i = j \\
0 & \text{otherwise}
\end{cases}
\end{align}

Thus, the optimization process not only pulls sample features closer to their centers but also dynamically updates each class's center to represent the mean of that class's feature distribution. This directly encourages features of the same class to aggregate together.

\subsection{Identity Density ($\text{ID}^2$) Metric}
Identity density is one aspect of measuring ReID effectiveness. However, there is currently no quantitative metric for this, and researchers commonly rely on visualization tools like t-SNE to demonstrate model performance. Due to the large number of IDs, this approach is limited to visualizing only a few IDs, making it challenging to assess model performance from a global perspective quantitatively. Some researchers exploit this limitation by selecting the best-performing IDs of their models for visualization. To address this, we propose an Identity Density ($\text{ID}^2$) Metric. This metric evaluates the global ID aggregation performance by taking each ID center across the entire test set (gallery and query) as a benchmark.
\begin{equation}
\text{ID}^2 = \frac{1}{N} \sum_{i=1}^{N} \frac{1}{n_i} \sum_{j=1}^{n_i} d\left( \frac{f_{ij}}{\|f_{ij}\|_2}, c_i \right)
\end{equation}
where \( N \) is the total number of unique IDs in the test set, and \( n_i \) is the number of samples for ID \( i \). The feature vector of the \( j \)-th sample of ID \( i \) is denoted as \( f_{ij} \), and \( c_i \) represents the identity center of ID \( i \), computed as follows:
\begin{equation}
c_i = \frac{1}{n_i} \sum_{j=1}^{n_i} \frac{f_{ij}}{\|f_{ij}\|_2}
\end{equation}

Both the feature vectors \( f_{ij} \) and the identity centers \( c_i \) are \( L_2 \)-normalized to ensure consistent feature scaling. The function \( d(\cdot, \cdot) \) represents the Euclidean distance.

\subsection{Pose Encoder Details}

The Pose Encoder module is designed to extract high-dimensional pose embeddings from the input poses. 
\begin{equation}
\mathbf{E}_{\text{pose}} = \text{PoseEncoder}(\mathbf{x}^{\text{pose}})
\end{equation}

The input is a feature map of size \(C_{\text{in}} \times H \times W\), denoted as \( \mathbf{x}^{\text{pose}} \), 
where \(C_{\text{in}}\) is the number of input channels, and \(H, W\) are the height, and width of the input. 
The first convolution layer is defined as:
\begin{equation}
\mathbf{E}_0 = \text{SiLU}(\text{Conv}_{\text{in}}(\mathbf{x}^{\text{pose}}))
\end{equation}
where \( \text{Conv}_{\text{in}} \) is a convolution operation with kernel size \(3 \times 3\), 
and the number of channels changes from \( C_{\text{in}} =3\) to \( C_0 =16\):

Each block applies a normal \(3 \times 3\) Conv, a \(3 \times 3\) Conv with stride 2 to reduce spatial dimensions, and followed by a SiLU activate function.
For the \(i\)-th convolutional block, the operations can be expressed as:
\begin{equation}
\mathbf{E}_{i+1} = \text{SiLU}(\text{Conv}_{i, \text{stride}=2}(\text{Conv}_{i}(\mathbf{E}_i)))
\end{equation}

The number of channels for each block is as follows: $[C_0, C_1, C_2, C_3] = [16, 32, 64, 128]$

The output Conv layer maps the features from the last block to the target embedding dimension \(C_{\text{out}} = 320\), 
expressed as:
\begin{equation}
\mathbf{E}_{\text{pose}} = \text{Conv}_{\text{out}}(\mathbf{E}_4)
\end{equation}

\subsection{Detailed Description of Neighbor Feature Centralization (NFC)}

\paragraph{Step 1: Compute Distance Matrix}

Given all feature vectors in the gallery \(\{\mathbf{z}_i\}_{i=1}^N\), our goal is to enhance each feature vector by aggregating features from its mutual nearest neighbors.
Compute the pairwise distance matrix \(\mathbf{D} = [d_{ij}]\) where \(d_{ij}\) represents the distance between features \(\mathbf{z}_i\) and \(\mathbf{z}_j\). To avoid self-matching, set the diagonal elements to a large constant, i.e. \[d_{ii} = C, \quad \text{for } i = 1, 2, \dots, N\]

\paragraph{Step 2: Find Top \(k_1\) Nearest Neighbors}

For each feature \(\mathbf{z}_i\), find its top \(k_1\) nearest neighbors based on the distance matrix \(\mathbf{D}\). Denote the set of indices of these neighbors as:
\begin{equation}
\mathcal{N}_{i} = \operatorname{TopK}_{k_1}(\{d_{ij}\}_{j=1}^N)
\end{equation}

\paragraph{Step 3: Identify Mutual Nearest Neighbors}

For each feature \(\mathbf{z}_i\), identifies its mutual nearest neighbors by checking whether each neighbor in \(\mathcal{N}_{i}\) also considers \(\mathbf{z}_i\) as one of its top \(k_2\) nearest neighbors. Specifically, for each \(j \in \mathcal{N}_{i}\), checks if \(i \in \mathcal{N}_{j}^{k_2}\), where \(\mathcal{N}_{j}^{k_2}\) is the set of indices of the top \(k_2\) nearest neighbors of \(\mathbf{z}_j\). If this condition is satisfied, add \(j\) to the mutual nearest neighbor set \(\mathcal{M}_{i}\):
\begin{equation}
\mathcal{M}_{i} = \{ j \mid j \in \mathcal{N}_{i}, \, i \in \mathcal{N}_{j}^{k_2} \}
\end{equation}

\paragraph{Step 4: Feature Centralization Enhancement}

Then it could centralize each feature vector \(\mathbf{z}_i\) by aggregating the features of its mutual nearest neighbors:
\begin{equation}
\mathbf{z}_i^{\text{centralized}} = \mathbf{z}_i + \sum_{j \in \mathcal{M}_{i}} \mathbf{z}_j
\end{equation}

This aggregation reduces feature noise and improves discriminability by incorporating information from similar features.

\begin{figure}
\centering
\includegraphics[width=\linewidth]{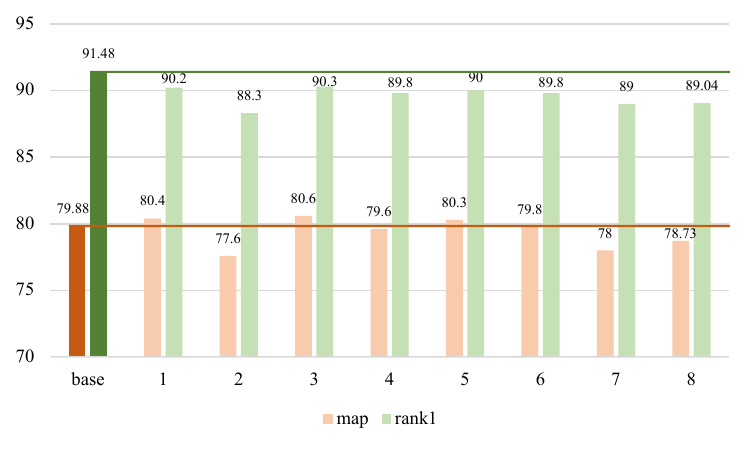}
\caption{ReID results with images generated with the same pose on Market1501.}
\label{fig:samepose}
\end{figure}

\begin{figure}
\centering
\includegraphics[width=0.9\linewidth]{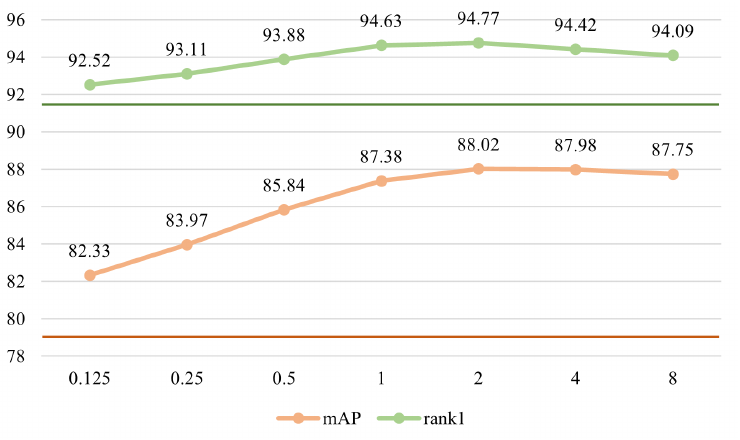}
\caption{Impact of the quality coefficient \( \eta \) with TransReID on Market1501. The dark color lines are the baseline.}
\label{fig:eta}
\end{figure}

\begin{figure}
\centering
\includegraphics[width=0.9\linewidth]{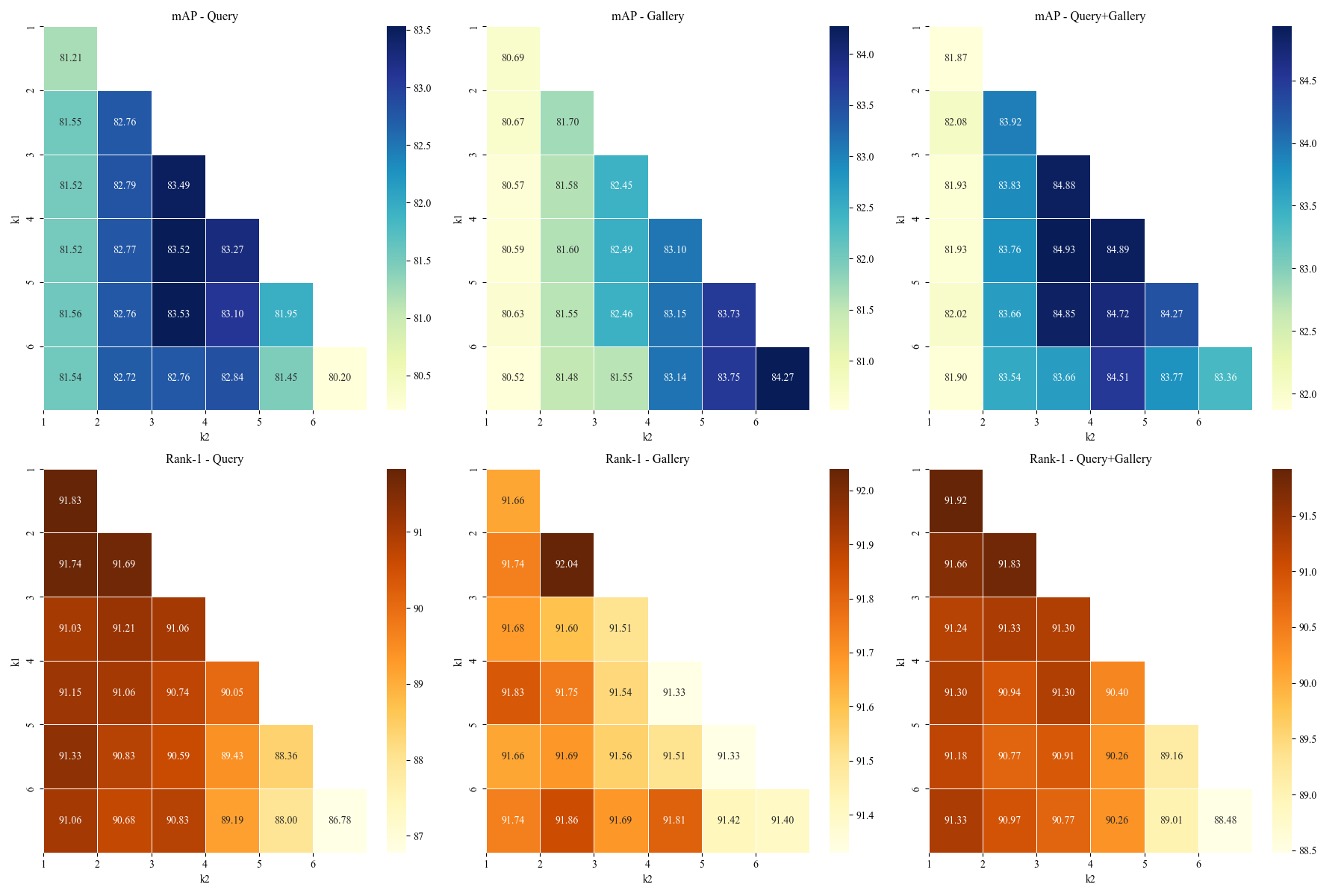}
\caption{$k_1/k_2$ analysis of Neighbor Feature Centralization (NFC) with TransReID on Market1501 without re-ranking.}
\label{fig:k1k2}
\end{figure}

\section{Experiments Supplementary}
\begin{figure*}
\centering
\includegraphics[width=\linewidth]{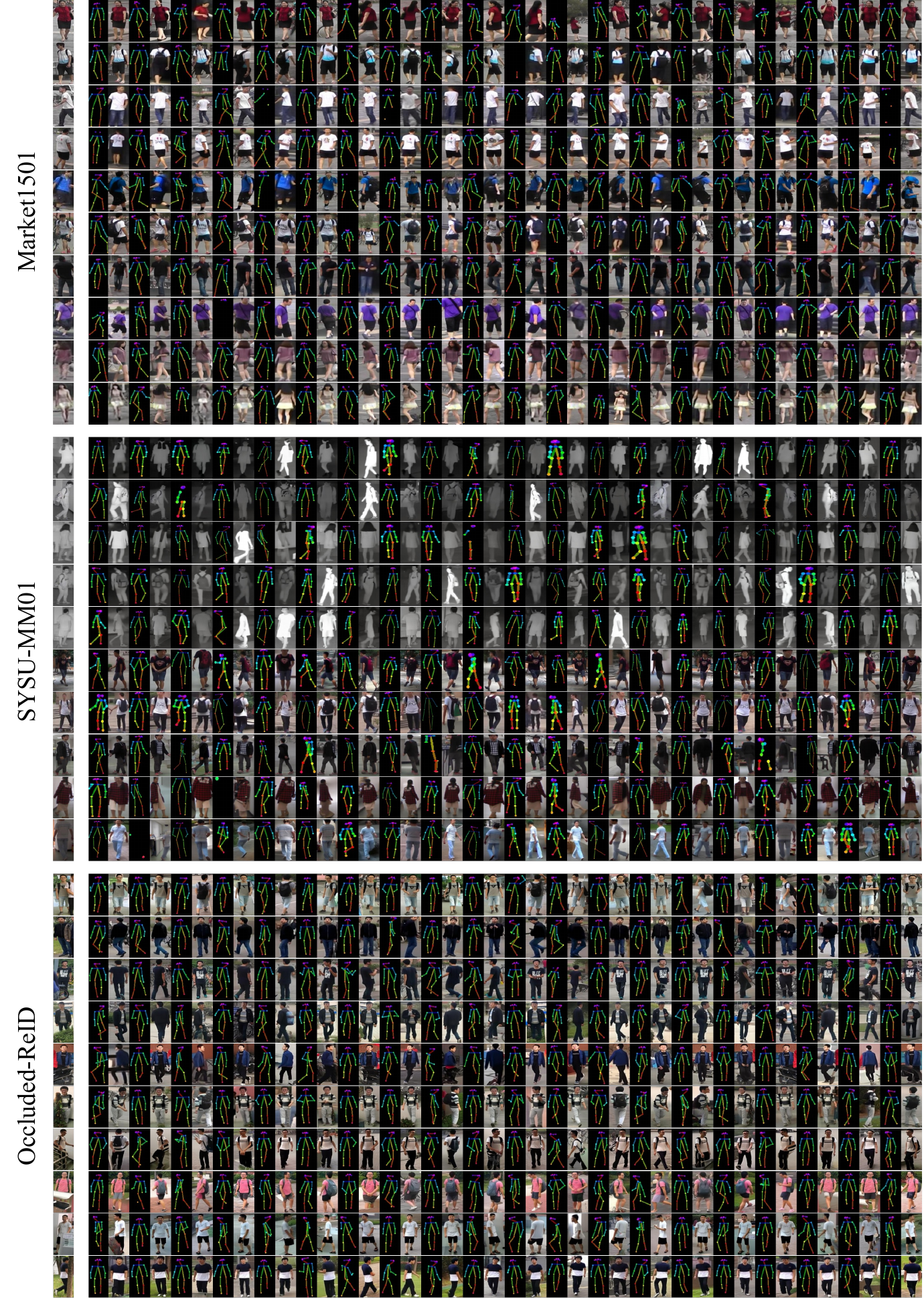}
\caption{More random generated images on three datasets.}
\label{fig:random_all}
\end{figure*}
\subsection{Data Cleansing}
Training an effective generative model requires high-quality data support. In current ReID (Person Re-Identification) datasets, there are many low-quality images, and removing them can help reduce interference to the model. In our experiments, we found two main issues that need to be addressed: \textbf{Extremely Low-quality Images}: The dataset contains images with such low resolution that even the human eye cannot recognize them as a "person". \textbf{Pose Estimation Failures}: The pose estimation model inevitably fails to detect pedestrian poses in some images.

\subsubsection{Extremely Low-quality Images}

To address this, manual filtering is impractical. Therefore, we designed an automated filtering algorithm. We leverage normal distribution of feature vector, if the feature on the edge of the distribution, largely due to the data itself is out of the distribution of its identity, and it can be picked up. 
% We calculate the mean and covariance matrix of the feature vectors for each ID and filter out samples whose feature distances lie outside a predefined quantile range.

Let \( \mathbf{f}_i \in \mathbb{R}^d \) denote the feature vector of the \( i \)-th sample of a particular identity, where \( d \) is the feature dimension. The mean vector \( \boldsymbol{\mu} \) and covariance matrix \( \boldsymbol{\Sigma} \) are computed as follows:
\begin{equation} 
\boldsymbol{\mu} = \frac{1}{N} \sum_{i=1}^{N} \mathbf{f}_i, \quad \boldsymbol{\Sigma} = \frac{1}{N} \sum_{i=1}^{N} (\mathbf{f}_i - \boldsymbol{\mu})(\mathbf{f}_i - \boldsymbol{\mu})^\top
\end{equation}
where \( N \) is the number of samples for a given ID.

To detect outliers, we compute the Mahalanobis distance \( d_i \) of each feature vector \( \mathbf{f}_i \) from the mean vector \( \boldsymbol{\mu} \), defined as:
\begin{equation}
dis_i = \sqrt{ (\mathbf{f}_i - \boldsymbol{\mu})^\top \boldsymbol{\Sigma}^{-1} (\mathbf{f}_i - \boldsymbol{\mu}) }
\end{equation}

Given that the feature vectors are assumed to follow a multivariate normal distribution, we use quantiles of the Mahalanobis distance to filter out outliers. Specifically, we define a lower bound \( Q_p \) and an upper bound \( Q_{1-p} \) based on the \( p \)-th and \( (1 - p) \)-th quantiles, respectively. Samples with distances outside this range are considered outliers and are removed, and can get a set $S^{\text{ref}}_{i}$ for $i^{th}$ ID:
\begin{equation} \label{outliers}
S_i^{\text{ref}} = \{ \mathbf{x}_i \mid dis_i \in [Q_p, Q_{1-p}] \}
\end{equation}

\begin{figure}
\centering
\includegraphics[width=0.8\linewidth]{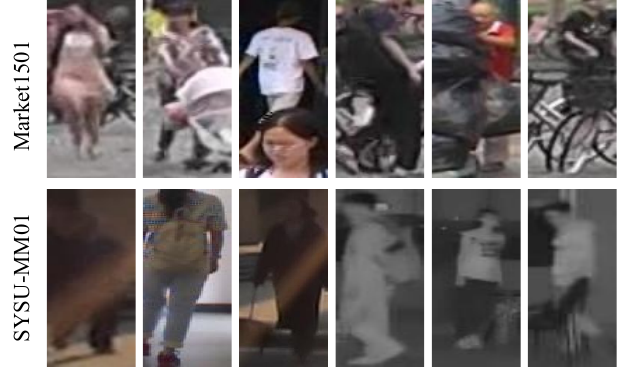}
\caption{Some outliers detected via the mechanism formulated as Equation.\ref{outliers} on Market1501 and SYSU-MM01 with quartile 0.005.}
\label{fig:outliers}
\end{figure}

\subsubsection{Pose Filtering for 'Failed' Pose Estimation}
\label{PoseValid}
We designed a pose filtering algorithm called PoseValid to eliminate cases where pose extraction has "completely failed." This algorithm checks the validity of the pose keypoints based on factors such as occlusion, keypoint positions, angles, and limb proportions, then get the set of valid poses.
\begin{equation}
S^{\text{trg}}_i = \{ \mathbf{x}_i \mid \text{PoseValid}(\mathbf{x}_i) \text{ and } dis_i \in [Q_p, Q_{1-p}] \}
\end{equation}
where the pose detector in this paper uses pretrained model of DWpose\cite{yang2023effective}. Given a set of keypoints representing a pose, we normalize the pose using the following steps:
\begin{enumerate}
    \item Compute the body height ($h$):\\
    Calculate the Euclidean distance between the Neck (keypoint 1) and the Left Hip (keypoint 11):
    \[
    h = \left\| \mathbf{k}_{\text{Neck}} - \mathbf{k}_{\text{LHip}} \right\|
    \]
    \item Translate the pose:\\
    Shift all keypoints so that the Neck is at the origin:
    \[
    \mathbf{k}'_i = \mathbf{k}_i - \mathbf{k}_{\text{Neck}}
    \]
    \item Scale the pose:\\
    Divide each keypoint by the body height to normalize the size:
    \[
    \mathbf{k}^{\text{normalized}}_i = \frac{\mathbf{k}'_i}{h}
    \]
\end{enumerate}
Then, the filtering process of PoseValid function evaluates the validity of pose keypoints by applying constraints on limb lengths, symmetry, and keypoint positions. 

\subsection{Generation quality and Pose Representation Study}
To assess the quality of the generated images, we replaced the real images in the dataset with images of the same pose and performed inference validation. The results, as shown in Fig.\ref{fig:samepose}, indicate that the original model still successfully matches pedestrians without significant performance degradation. Even with all images in the same pose, the model can effectively differentiate between individuals. This suggests that our generated images are of high quality, retaining the main characteristics of the original images without notably impacting the ReID model. Moreover, we found that pedestrians walking at an angle have higher distinguishability compared to other poses (front, back, and side views), which are more representative of their identities.

\begin{table}[]
\small
\renewcommand{\arraystretch}{1.0}
\renewcommand\tabcolsep{5pt}
\centering
    \begin{tabular}{cc|ccc}
        \hline
        +Ours & +Rerank & mAP & Rank1  \\ \hline
        \ding{55} &\ding{55} & 79.88 & 91.48 \\
        \ding{55} &\ding{51} & 89.56 & 92.07  \\ \hline
        \rowcolor{gray!20}
        \ding{51} &\ding{55} &90.39&94.74\\
        \rowcolor{gray!20}
        \ding{51} &\ding{51}  & \textbf{92.79}&	\textbf{94.83} \\ \hline
    \end{tabular}
    \caption{Compared to k-reciprocal rerank with official settings on Market1501 ($k_1$=20,$k_2$=6).}
    \label{tab:rerank}
\end{table}

\begin{table}[]
\small
\renewcommand{\arraystretch}{1.0}
\renewcommand\tabcolsep{5pt}
\centering
        \begin{tabular}{c|ccc}
        \hline
        Methods & mAP & Rank1  \\ \hline
        TransReID on MSMT17 & 67.80	&85.33\\
        \rowcolor{gray!20}
        +ours & 74.06	&86.55	  \\ \hline
    \end{tabular}
    \caption{Experiment on MSMT17 with TransReID and their official weights.}
    \label{tab:more}
\end{table}

\subsection{More Random Generation}
We provide additional randomly generated images in Fig.\ref{fig:random_all} from Market-1501, SYSU-MM01 and Occluded-ReID datasets.

\subsection{Collaborate with Re-ranking}
Since our method does not change the features' original distribution,
it could collaborate post-processing strategies like rerank, as shown in Tab.\ref{tab:rerank}.

\subsection{Results on MSMT17 with TransReID}
We conduct a simple experiment on MSMT17 dataset with with TransReID and their official pre-trained weights. As shown in Tab.\ref{tab:more}.

\subsection{Comparisons with state-of-the-art methods on three ReID benchmarks}
Comparison on three ReID benchmarks. Since Our method can be applied to any baseline, we choose three methods from three benchmarks which have the official codes and pre-trained weights. With our method, we achieve the new SOTA in three benchmarks, as shown in Fig.\ref{tab:sota_m} and Fig.\ref{tab:sys}.

\begin{table}
\centering
\renewcommand\tabcolsep{5pt}

\begin{tabular}{c|cc|cc}
\hline
 \multirow{2}{*}{Methods}& \multicolumn{2}{c|}{Market1501} & \multicolumn{2}{c}{Occluded-reID} \\ \cline{2-5}
 & Rank-1 & mAP & Rank-1  & mAP \\
\hline
BoT\cite{luo2019bag} & 94.5 & 85.9 & 58.4 & 52.3 \\
PCB\cite{sun2018beyond}& 93.8 & 81.6& - & - \\
VGTri\cite{yang2021learning} & - & - & 81.0 & 71.0 \\
PVPM\cite{gao2020pose} & - & - & 66.8 & 59.5 \\
HOReID\cite{wang2020high} & 94.2 & 84.9 & 80.3 & 70.2 \\
ISP\cite{zhu2020identity} & 95.3 & 88.6 & - & -\\
PAT\cite{li2021diverse} & 95.4 & 88.0 & 81.6 & 72.1 \\
TRANS\cite{he2021transreid} & 95.2 & 88.9 & - & - \\
CLIP\cite{li2023clip} & 95.7 & 89.8 & - & - \\
SOLIDER\cite{chen2023beyond} & 96.9 & 93.9 & - & - \\
SSGR\cite{yan2021occluded} & 96.1 & 89.3 & 78.5 & 72.9 \\
FED\cite{wang2022feature} & 95.0 & 86.3& 86.3 & 79.3 \\
BPBreid\cite{somers2023body} & 95.7 & 89.4 & 82.9 & 75.2\\
PFD\cite{wang2022pose} & 95.5 & 89.7 & 83.0 & 81.5 \\
KPR\textsubscript{IN}\cite{somers2025keypoint} & 95.9 & 89.6 & 85.4 & 79.1 \\
KPR\textsubscript{SOL}\cite{somers2025keypoint} & 96.62 &93.22 &84.83& 82.6\\
\hline
\rowcolor{gray!20}
CLIP+ours & 97.3 & 94.9 & - & -\\
\rowcolor{gray!20}
KPR\textsubscript{IN}+ours & - & - & 91 & 89.34 \\
\hline
\end{tabular}
\caption{Comparisons with state-of-the-art methods on Market1501 and Occluded-reID.}
\label{tab:sota_m}
\end{table}
\begin{table}
\small
\centering
\renewcommand\tabcolsep{5pt}
	\begin{tabular}{l|cc|cc}
		\hline
		\multirow{2}{*}{Methods} & \multicolumn{2}{c|}{All-Search} &\multicolumn{2}{c}{Indoor-Search}\\ \cline{2-5}
              & mAP & Rank-1 & mAP & Rank-1 \\ \cline{1-5}
             PMT\cite{lu2023learning}& 66.13& 67.70& 77.81& 72.95\\
             MCLNet~\cite{hao2021cross}& 61.98 &65.40&76.58 &72.56\\
             MAUM~\cite{Liu2022LearningMU}& 68.79 &71.68& 81.94 &76.9\\
             CAL\cite{CAL}& 71.73 &74.66& 83.68 &79.69\\
             SAAI(w/o AIM)~\cite{fang2023visible}&  71.81& 75.29&  84.6 &81.59\\
             SEFL\cite{feng2023shape}& 72.33 &77.12& 82.95 &82.07\\
             PartMix\cite{kim2023partmix}& 74.62 &77.78& 84.38 &81.52\\
             MID~\cite{Huang2022ModalityAdaptiveMA} & 59.40 &60.27& 70.12 &64.86\\
             FMCNet~\cite{zhang2022fmcnet}& 62.51 &66.34& 74.09 &68.15\\
             MPANet~\cite{wu2021discover}& 68.24 &70.58& 80.95 &76.74\\
             CMT~\cite{jiang2022cross}& 68.57 &71.88& 79.91 &76.90\\
             protoHPE~\cite{zhang2023protohpe}& 70.59 &71.92&81.31 &77.81\\
             MUN~\cite{yu2023modality}& 73.81 &76.24& 82.06 &79.42\\
             MSCLNet~\cite{zhang2022modality}& 71.64 &76.99& 81.17 &78.49\\
             DEEN~\cite{zhang2023diverse}& 71.80 &74.70& 83.30 &80.30\\
             CIFT~\cite{li2022counterfactual}& 74.79 &74.08& 85.61 &81.82\\
             \cline{1-5}
             \rowcolor{gray!20}
            SAAI+ours&76.44&79.33& 86.83& 84.2\\
             \cline{1-5}
	\end{tabular}
             \caption{Comparison with state-of-the-art methods on SYSU-MM01 without re-ranking.}
             \label{tab:sys}                               
\end{table}

% \begin{figure*}
% \centering
% \includegraphics[width=\linewidth]{figs/pdf/random.pdf}
% \caption{Random generated images on Market1501.}
% \label{fig:random_market}
% \end{figure*}

% \begin{figure*}
% \centering
% \includegraphics[width=\linewidth]{figs/pdf/sysu.pdf}
% \caption{Random generated images on SYSU-MM01.}
% \label{fig:random_sysu}
% \end{figure*}

% \begin{figure*}
% \centering
% \includegraphics[width=\linewidth]{figs/pdf/occ.pdf}
% \caption{Random generated images on SYSU-MM01.}
% \label{fig:random_occ}
% \end{figure*}

\subsection{Analysis on quality coefficient $\eta$ of Generation Model}

Fig.\ref{fig:eta} illustrates the effect of adjusting the coefficient $\eta$ on the performance of the ReID model. To evaluate this impact, we gradually increased the value of $\eta$ and observed changes on the mAP and Rank-1 metrics. 

As the value of $\eta$ increases, the performance of the ReID model improves, reaching an optimal point. At $\eta = 2$, both mAP and Rank-1 achieve their maximum values of 88.02\% and 94.77\%, respectively. However, further increasing $\eta$ beyond this point leads to a slight decline in performance. It is easy to find that using generated images to centralize features is effective. However, considering the quality of the generated image, direct adding, although also effective, may not always achieve the best results. Therefore adjusting $\eta$ according to the generation quality of the model in this dataset can better centralize the features.

\subsection{Analysis on $k_1/k_2$ of Neighbor Feature Centralization}
We conducted a detailed analysis of different $k_1$ and $k_2$ combinations, evaluating the results of feature centralization enhancement separately on the Query and Gallery sets, as well as the combined effect (as shown in the Fig.\ref{fig:k1k2}). The selection of these two parameters primarily depends on the number of potential positive samples within the set (adjusting $k_1$) and the confidence in feature associations (adjusting k2). Overall, medium parameter combinations ($k_1$ and $k_2$ in the range of 2-4) provide relatively optimal performance.

\end{document}